# Sokoto Coventry Fingerprint Dataset

**Yahaya Isah Shehu**[1✉], **Ariel Ruiz-Garcia**[1✉], **Vasile Palade**[1] **and Anne James**[2]

1 Coventry University, Faculty of Engineering, Environment and Computing.
Priory Street, CV1 5FB, Coventry, UK
[2] Nottingham Trent University, Faculty of Science and Technology,
Clifton Campus, NG11 8NS, Nottingham, UK
`shehuy2,ariel.ruiz-garcia,vasile.palade@coventry.ac.uk,`
`anne.james@ntu.ac.uk`

**Abstract.** This paper presents the Sokoto Coventry Fingerprint Dataset (SOCO-Fing), a biometric fingerprint database designed for academic research purposes. SOCOFing is made up of 6,000 fingerprint images from 600 African subjects. SOCOFing contains unique attributes such as labels for gender, hand and finger name as well as synthetically altered versions with three different levels of alteration for obliteration, central rotation, and z-cut. The dataset is freely available for noncommercial research purposes at: `https://www.kaggle.com/ruizgara/socofing`

## 1 Introduction

The SOCOFing dataset contains 6,000 fingerprints belonging to 600 African subjects. There are 10 fingerprints per subject and all subjects are 18 years or older. SOCOFing contains unique attributes such as labels for gender, hand and finger name. Moreover, synthetically altered versions of these fingerprints are provided with three different levels of alteration for obliteration, central rotation, and z-cut using the STRANGE toolbox [1]. STRANGE is a novel framework for the generation of realistic synthetic alterations on fingerprint images [1]. Alterations were done using easy, medium and hard parameter settings in the STRANGE toolbox over 500dbi resolution images. Therefore we provide a total of 17,934 altered images with easy parameter settings, 17,067 with medium settings, and 14,272 with hard parameter settings. Note that in some cases some images did not meet the criteria for alteration with specific settings using the STRANGE toolbox, hence the unequal number of altered images across all three alteration categories.

All original images were acquired based on impressions collected with Hamster plus (HSDU03P[TM]) and SecuGen SDU03P[TM] sensor scanners. SOCOFing consists of a total of 55,273 fingerprint images combined. All file images have a resolution of $1 \times 96 \times 103$ (gray $\times$ width $\times$ height). Figure 2 below shows some samples of the original fingerprints.

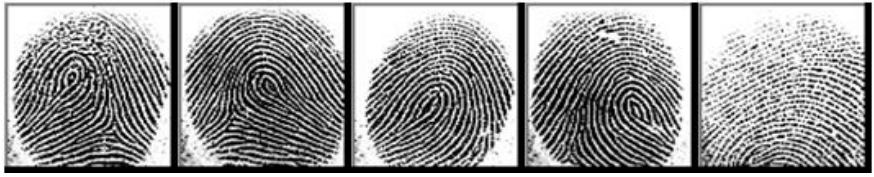

**Fig. 1.** Sample illustration of five left hand fingerprints belonging to the same subject.

Figure 2 below shows the synthetic alteration and generation of the fingerprints images into Z-cut, obliteration and central rotation.

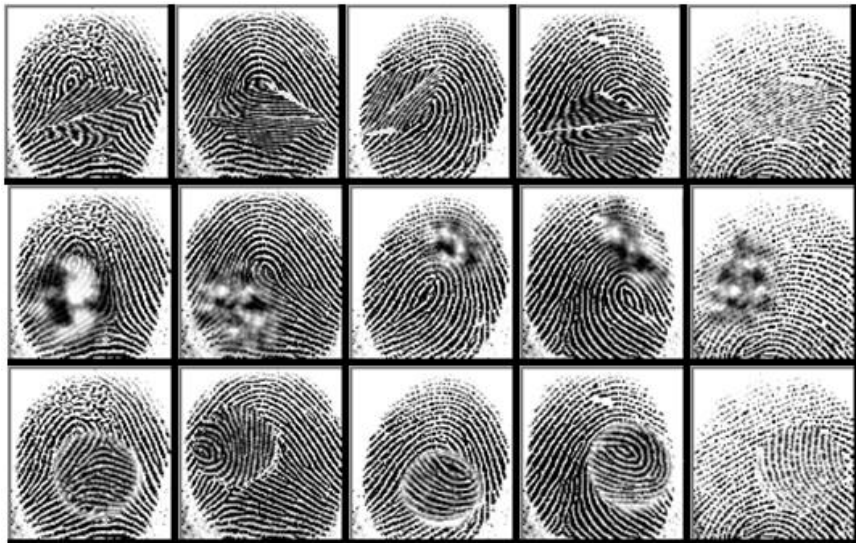

**Fig. 2.** Images from Figure 1 after being altered into z-cut, obliteration and central rotation.

## 2    Nomenclature

The dataset is divided into two subfolders containing real, i.e. original images, and altered images. The altered folder is further divided into three levels of alteration difficulty: easy, medium and hard.

$$\text{SOCOFing} \begin{cases} \text{Real} \\ \text{Altered} \begin{cases} \text{Altered-Easy} \\ \text{Altered-Medium} \\ \text{Altered-hard} \end{cases} \end{cases}$$

The file format provides the labels for each individual image and has the naming convention of:

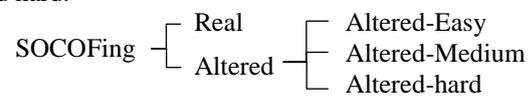

"001_M_Left_little_finger_Obl.bmp"
 1   2   3      4          5   6

where:

1. Identifies the number of the subject: 001 to 600.
2. Indicates the gender of the subject: M – male, F – female.
3. Denotes the hand: Left or Right.
4. Indicates the finger name: little, ring, middle, index, or thumb.
5. Indicates the type of alteration type (altered images only): Obl – obliteration, CR – central rotation, or Zcut.
6. File extension: ".bmp" for all images.

## 3    Usage

This dataset is provided AS IS for noncommercial not-for-profit research purposes only. Any publications arising from the use of this software, including but not limited to academic journal and conference publications, technical reports and manuals, must cite this document and the following work:

*Shehu, Y.I., Ruiz-Garcia, A., Palade, V., James, A. (2018) "Detection of Fingerprint Alterations Using Deep Convolutional Neural Networks" in Proceedings of the International Conference on Artificial Neural Networks (ICANN 2018), Rhodes – Greece, 5$^{th}$ - 7th October 2018. Springer-Verlag Lecture Notes in Computer Science.*

For any questions related to the dataset or the work referred above please contact the corresponding authors.